%% file: main.tex
\pdfoutput=1
\documentclass[letterpaper]{article} % DO NOT CHANGE THIS
\usepackage{aaai2026}  % DO NOT CHANGE THIS
\usepackage{times}  % DO NOT CHANGE THIS
\usepackage{helvet}  % DO NOT CHANGE THIS
\usepackage{courier}  % DO NOT CHANGE THIS
\usepackage[hyphens]{url}  % DO NOT CHANGE THIS
\usepackage{graphicx} % DO NOT CHANGE THIS
\usepackage{natbib}  % DO NOT CHANGE THIS AND DO NOT ADD ANY OPTIONS TO IT
\usepackage{caption} % DO NOT CHANGE THIS AND DO NOT ADD ANY OPTIONS TO IT
\usepackage{algorithm}
\usepackage{algorithmic}

\usepackage{newfloat}
\usepackage{listings}
\DeclareCaptionStyle{ruled}{labelfont=normalfont,labelsep=colon,strut=off} % DO NOT CHANGE THIS
\floatstyle{ruled}
\newfloat{listing}{tb}{lst}{}
\floatname{listing}{Listing}
\usepackage{xcolor}         % colors
\usepackage{colortbl}

\definecolor{mygreen}{rgb}{0.1294,0.6980,0.6706}
\definecolor{myblue}{rgb}{0.2784,0.6784,1}
\usepackage{multirow}
\usepackage{algorithm}
\usepackage{nicematrix}
\usepackage{amssymb}

\usepackage{subfig}
\usepackage{booktabs} 

\nocopyright 
\title{APFL: Analytic Personalized Federated Learning via Dual-Stream Least Squares}
\author{
    Kejia Fan\textsuperscript{\rm 1},
    Jianheng Tang\textsuperscript{\rm 2},
    Zhirui Yang\textsuperscript{\rm 1},
    Feijiang Han\textsuperscript{\rm 3},
    Jiaxu Li\textsuperscript{\rm 1},
    Run He\textsuperscript{\rm 4},
    \\
    Yajiang Huang\textsuperscript{\rm 1},
    Anfeng Liu\textsuperscript{\rm 1},
    Houbing Herbert Song\textsuperscript{\rm 5},
    Yunhuai Liu\textsuperscript{\rm 2},
    Huiping Zhuang\textsuperscript{\rm 4}
}
\affiliations{
    \textsuperscript{\rm 1} CSU, China \quad  \textsuperscript{\rm 2} pku, China  \quad  \textsuperscript{\rm 3} UPenn, USA \quad \textsuperscript{\rm 4} SCUT, China \quad \textsuperscript{\rm 5} UMBC, USA\\
}

\usepackage{bibentry}
\begin{document}

\maketitle

\input{sec/0_abstract}    

\input{sec/1_intro}
\input{sec/2_related}
\input{sec/3_method}
\input{sec/4_experiments}

\bibliography{main}
\end{document}

%% file: sec/0_abstract.tex
\begin{abstract}
Personalized Federated Learning (PFL) has presented a significant challenge to deliver personalized models to individual clients through collaborative training.
Existing PFL methods are often vulnerable to non-IID data, which severely hinders collective generalization and then compromises the subsequent personalization efforts.
In this paper, to address this non-IID issue in PFL, we propose an \underline{A}nalytic \underline{P}ersonalized \underline{F}ederated \underline{L}earning (APFL) approach via dual-stream least squares.
In our APFL, we use a foundation model as a frozen backbone for feature extraction.
Subsequent to the feature extractor, we develop dual-stream analytic models to achieve both collective generalization and individual personalization.
Specifically, our APFL incorporates a shared primary stream for global generalization across all clients, and a dedicated refinement stream for local personalization of each individual client.
The analytical solutions of our APFL enable its ideal property of heterogeneity invariance, theoretically meaning that each personalized model remains identical regardless of how heterogeneous the data are distributed across all other clients.
Empirical results across various datasets also validate the superiority of our APFL over state-of-the-art baselines, with advantages of at least 1.10\%-15.45\% in accuracy.
\end{abstract}

%% file: sec/1_intro.tex
\section{Introduction}
\label{sec:intro}

Federated Learning (FL) has emerged as a prominent distributed machine learning paradigm, enabling the collaborative training of a global model across numerous clients while safeguarding individual data privacy~\cite{FL-1,FL-2,FL-3}.
Yet, a single, universally shared model of traditional FL often falls short in accommodating the diverse, personalized requirements of each client's local applications~\cite{PFL-0,PFL-1}. 
In this context, Personalized Federated Learning (PFL) has come in and gained prominence as an extension of the FL paradigm, aiming to construct personalized models for each client in addition to the global model learned collaboratively among all clients.
PFL offers a dual promise for both generalization and personalization~\cite{PFL-5}.
First, it enables each local client to benefit from the collective knowledge shared by all participants in the federation, thereby enhancing the model's overall generalization capabilities~\cite{PFL-4}. 
Second, PFL also facilitates individual personalization, allowing models to adapt specifically to the unique characteristics of each client's local data~\cite{PFL-3}.

\begin{figure}[t]
\centering
\includegraphics[width=0.965\columnwidth]{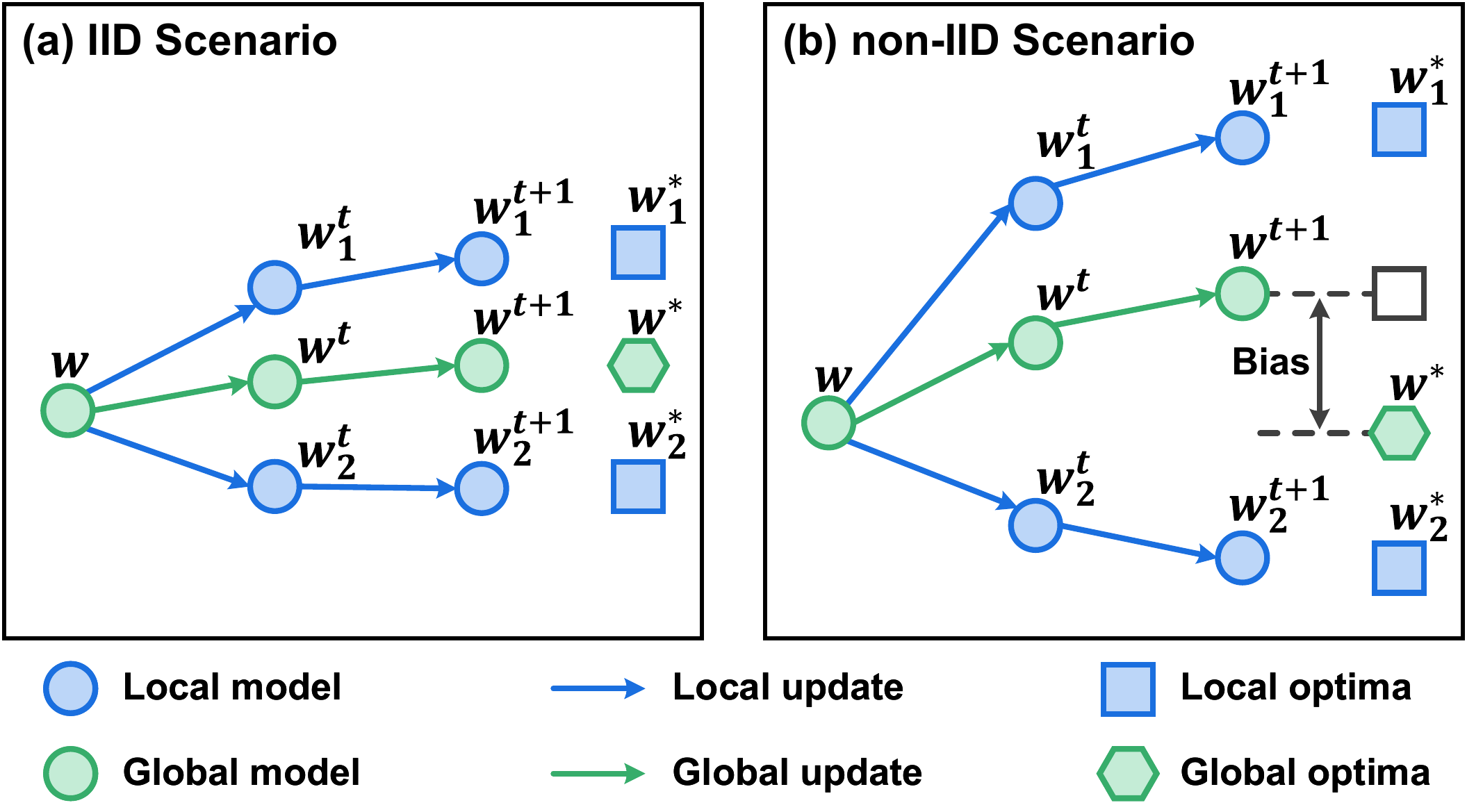}
\caption{Data heterogeneity in PFL.}
\label{Figure-1}
% \vspace{4.2em}
\end{figure}

Data heterogeneity, often referred to as non-Independent and Identically Distributed (non-IID) data, represents a key challenge within PFL~\cite{PFL-0,PFL-1}. 
As depicted in Fig.~\ref{Figure-1}, in a non-IID environment, the local gradients of client-side training are naturally biased towards their unique local data distributions.
Such bias leads to distinct local optima and conflicting learning directions among clients.
Consequently, the diverse local datasets can cause the aggregated global model to drift away from a truly generalized representation, degrading its overall performance~\cite{PFL-2}.
Existing PFL methods are typically susceptible to this issue, which severely hinders collective generalization and then compromises the subsequent personalization efforts.

This observation reminds us: \textbf{\textit{the core challenge in PFL is rooted in the long-standing reliance on gradient-based updates, which are inherently vulnerable to non-IID data}}.
This insight has already been recognized in many existing studies~\cite{PFL-0,PFL-3}.
Thus, a promising and natural avenue to address the gradient-related concern may be eliminating the use of gradient-based updates.
As a prominent gradient-free technique~\cite{GACL}, analytic learning has exhibited great promise and superiority within FL~\cite{AFL}.
However, existing Analytic Federated Learning (AFL) approach is limited to traditional FL with a single universal model, falling short in the prevalent PFL scenarios, where each client has a unique personalized requirement~\cite{AFL}.
Moreover, AFL directly performs linear classification on the backbone's output features, introducing a potential risk of under-fitting~\cite{AFL}.

In this paper, to address the non-IID issue in PFL, we propose an \underline{A}nalytic \underline{P}ersonalized \underline{F}ederated \underline{L}earning (APFL) approach to address the non-IID issue in PFL.
In our APFL, we use a foundation model as a frozen backbone for feature extraction.
Subsequent to the feature extractor, we develop dual-stream analytic models.
Specifically, our APFL incorporates a shared primary stream for global generalization across all clients, and a dedicated refinement stream for local personalization of each individual client.
To enhance the fitting ability of our APFL, we employ random projections and nonlinear activations to improve the feature's linear separability.
In addition, our APFL supports different activated features within the dual streams to capture collective generalization and individual personalization in distinct feature spaces.
Our key contributions are summarized as follows.
\begin{enumerate}
    \item To our knowledge, our APFL presents the first endeavor to bring analytic learning into PFL for handling the issue of non-IID data via dual-stream least squares.
    \item We develop dual analytic streams of least squares from distinct activated features to achieve both collective generalization and individual personalization in PFL.
    \item We conduct theoretical analyses to establish our APFL's ideal property of heterogeneity invariance, i.e., each personalized model remains identical regardless of how non-IID the data are distributed across all other clients.
    \item We conduct experiments across various datasets to show the superiority of our APFL, outperforming competitive baselines by at least 1.10\%-15.45\% in accuracy.
\end{enumerate}

%% file: sec/2_related.tex
\section{Related Work}
\label{sec:related}

\subsection{Personalized Federated Learning}
FL has emerged as a prominent distributed machine learning paradigm, allowing clients to collaboratively train a shared global model~\cite{FL-1,FL-3}. 
However, this ``one-size-fits-all" style of traditional FL often falls short when clients have diverse data distributions and unique application requirements~\cite{PFL-0,PFL-5}.
In this context, PFL has recently gained promise to extend the traditional FL, aiming to collaboratively train models that are not only globally informed but also individually tailored to each client's requirement~\cite{PFL-3}.
PFL's core appeal lies in its dual promise for both collective generalization and individual personalization~\cite{PFL-4,PFL-1}.
Despite the promising prospects offered by PFL, a critical accompanying challenge is the presence of non-IID data across clients.
This data heterogeneity severely hinders convergence and degrades model performance by impeding the global aggregation.
It is widely recognized that this challenge within PFL stems from the vulnerability of gradient-based updates to non-IID data~\cite{PFL-0}.
Although numerous efforts are underway to address the issue of non-IID data in PFL, most approaches focus on mitigating symptoms rather than confronting the root cause within the gradients~\cite{PFL-0,PFL-1}.
Differing from prior research in PFL, our proposed APFL aims to fundamentally avoid gradient-based updates through analytical (i.e., closed-form) solutions, enabling it to achieve the ideal and rare property of heterogeneity invariance.

\subsection{Analytic Learning}
Analytic learning presents a gradient-free alternative to traditional backpropagation to circumvent gradient-related issues such as vanishing and exploding gradients~\cite{AL_1}.
This technique is also referred to as pseudoinverse learning due to its reliance on matrix inversion~\cite{AL_2, CALM}.
Its core idea is to directly derive analytical (i.e., closed-form) solutions using least squares~\cite{TS-ACL,CFSSeg}. 
To ease the memory demands inherent in analytic learning, researchers propose the block-wise recursive Moore-Penrose inverse to enable efficient joint learning~\cite{AL_6}.
Thanks to this breakthrough, analytic learning has shown superiority in modern applications, particularly in continual learning~\cite{GACL,ACU,AFCL}.
While AFL has been proposed to introduce analytic learning into FL, it is limited to traditional FL with a single universal model, falling short in the PFL scenarios~\cite{AFL}.
Our APFL bridges this gap via dual-stream least squares, achieving both generalization and personalization in PFL.
In addition, to enhance the fitting capabilities of analytic learning, we also improve the linear separability of features by incorporating random projections and nonlinear activations.

%% file: sec/3_method.tex
\section{Our Proposed APFL}
\label{sec:method}

We consider the standard PFL scenario with one server and $K$ clients.
Each client $k$ possesses its local dataset $D_k = \{\mathbf{X}_k, \mathbf{Y}_k\}$, where $\mathbf{X}_k \in \mathbb{R}^{N_k \times l\times w\times h}$ and $\mathbf{Y}_k \in \mathbb{R}^{N_k \times c}$ represent the $N_k$ local data samples and their corresponding labels, respectively.
Here, $l\times w\times h$ denotes the 3 dimensions of input images, and $c$ denotes the number of classes.
On the one hand, the clients aim to improve their models' generalization capabilities by leveraging the collective knowledge within the federation.
On the other hand, the clients seek to enhance their models' personalization capabilities by adapting to the unique characteristics of their own local data.

\subsection{Motivation and Overview}

\begin{figure*}[t]
\centering
\includegraphics[width=2.0\columnwidth]{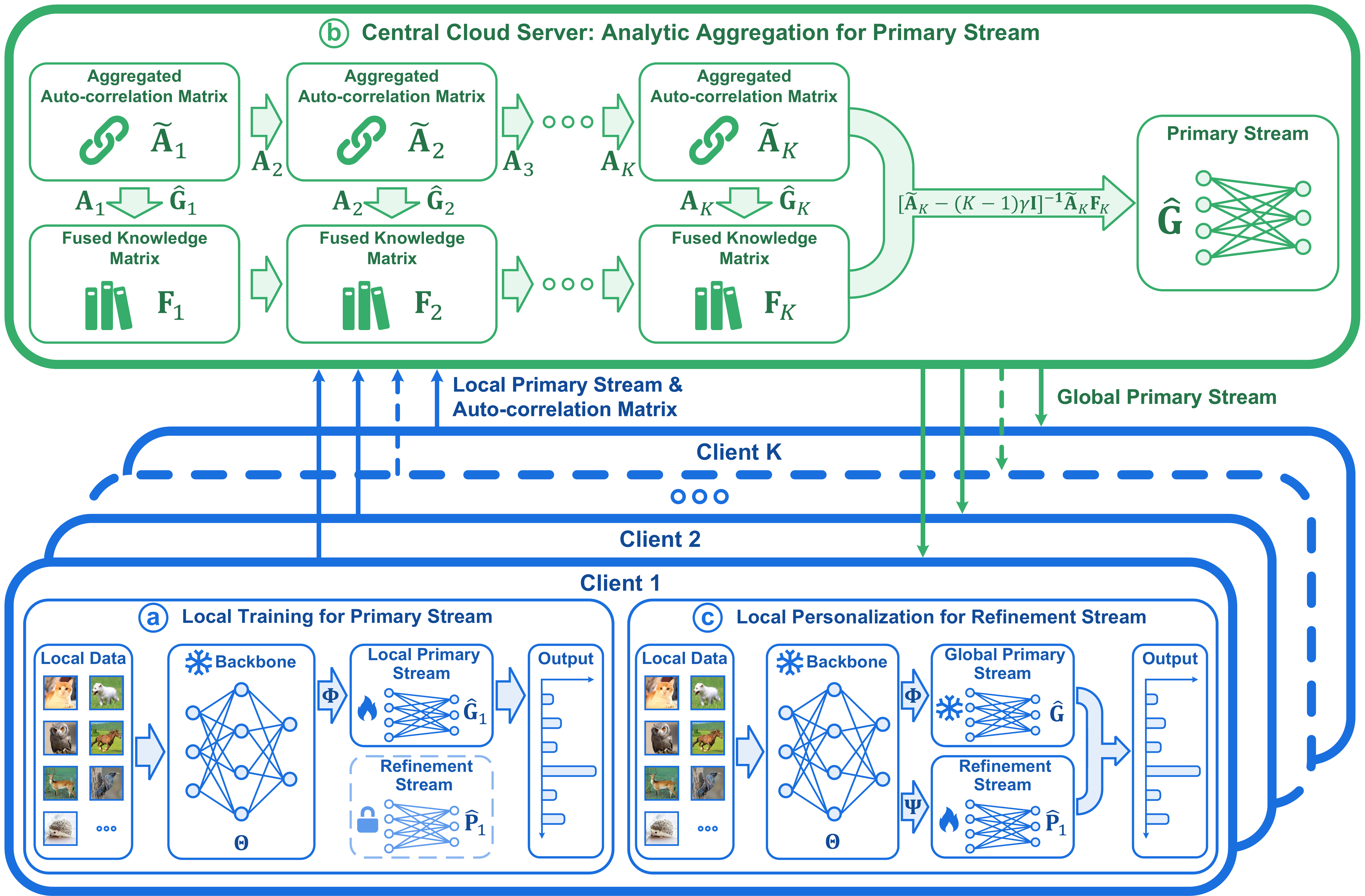}
\caption{Detailed design of our proposed APFL.}
\label{fig_2}
\end{figure*}

As previously analyzed, the core challenge in PFL is rooted in its reliance on gradient-based updates, rendering models inherently vulnerable to non-IID data.
Motivated by the emergence of analytic learning as a promising gradient-free alternative, we are the first to integrate this methodology into PFL, thereby fundamentally avoiding the reliance on gradient-based updates. 
Considering the dual requirements of generalization and personalization in PFL, we develop dual-stream analytic models as illustrated in Figure~\ref{fig_2}.

Specifically, capitalizing on the exceptional performance of pre-trained models and their widespread adoption in PFL \cite{PFL-backbone-1,PFL-backbone-2,PFL-backbone-3}, we first employ a foundation model, e.g., ViT~\cite{ViT-MAE}, as the frozen backbone to obtain powerful feature representations.
Following this, we develop the primary stream $\mathbf{\hat G}$ to aggregate collective knowledge from all $K$ clients within the federation, thereby enhancing global generalization.
Finally, for each client $k$, we further construct the individual refinement stream $\mathbf{\hat P}_k$ to capture and adapt to the client’s unique local preferences, thereby enabling local personalization.

Thus, our goal is to obtain a primary stream $\mathbf{\hat G}$ and a total of $K$ refinement streams $\{\mathbf{\hat P}_k\}_{k=1}^{K}$, as formulated below.

\noindent\textbf{Minimize:}
\begin{equation}
\sum\nolimits_{k=1}^K\|(\mathbf{Y}_{k}-\mathbf{\Phi}_{k}\mathbf{\hat G})-\mathbf{\Psi}_{k}\mathbf{\hat P}_k\|^2_\text{F}+{\color{black}\mathbf{\beta} \| \mathbf{\hat P}_k \|^{2}_\text{F}},
\label{eq:objective-1}
\end{equation}
\noindent\textbf{Subject to:}
\begin{equation}
\mathbf{\hat G} = \arg\underset{\mathbf{G}}{\min} \ 
  {\color{black}\|\mathbf{Y}_{1:K}-\mathbf{\Phi}_{1:K}\mathbf{G}\|^2_\text{F}}+{\color{black}\mathbf{\gamma} \| \mathbf{G} \|^{2}_\text{F}}.
  \label{eq:objective-2}
\end{equation}

Here, $\mathbf{\Phi}_{k}$ and $\mathbf{\Psi}_{k}$ represent the client $k$'s activated feature matrices, both derived from the frozen backbone yet subsequently passed through distinct random projections and nonlinear activation functions.
The objective function \eqref{eq:objective-1} serves as the standard PFL objective, which minimizes the sum of each client's local empirical risk for local personalization.
The constraint \eqref{eq:objective-2} ensures that the primary stream minimizes the global empirical risk for global generalization.

Notably, since the cross‐entropy loss does not admit a closed‐form solution, we adopt the Mean Squared Error (MSE) as the loss.
In fact, the MSE loss is widely used in analytic learning \cite{GACL,AFL} and has been shown to achieve performance comparable to that of cross-entropy \cite{MSE-1}.
Subsequently, leveraging dual-stream least squares, we devise the corresponding computation process for PFL to analytically derive the optimal solutions to \eqref{eq:objective-2} and \eqref{eq:objective-1}, thereby achieving excellent performance in both generalization and personalization.

\subsection{Primary Stream for Global Generalization}

Initially, all clients collaboratively train a shared primary stream for global generalization. 
To achieve this objective, we meticulously design the computation and aggregation process to derive the closed-form solution to \eqref{eq:objective-2} analytically, rendering our approach immune to non-IID data.

Specifically, each client $k$ first uses the established backbone of the foundation model for feature extraction, thereby obtaining its primary-stream feature matrix $\mathbf{\Phi}_{k}$ via \eqref{eq:section-3-2-1}.
\begin{equation}
    \mathbf{\Phi}_{k} = \sigma_\text{P} (\text{Backbone}(\mathbf{X}_k, \Theta) \mathbf{R}_\text{P}), \label{eq:section-3-2-1}
\end{equation}
where $\mathbf{\Phi}_{k} \in \mathbb{R}^{N_k \times d_\text{P}}$ denote the activated feature matrix of client $k$ and $d_\text{P}$ is the corresponding feature dimension.
Meanwhile, $\text{Backbone}(\, \cdot \,, \Theta)$ represents the backbone with its parameters $\Theta$, while $\mathbf{R}_\text{P}$ and $\sigma_\text{P} (\cdot)$ represent the random projection matrix and the nonlinear activation function.

Subsequently, the client utilizes its local activated feature $\mathbf{\Phi}_{k}$ and label $\mathbf{Y}_{k}$ to construct its local primary stream $\mathbf{\hat G}_k$. 
Specifically, the problem of training the local primary stream can be formulated as the following ridge regression problem:
\begin{equation}
\begin{aligned}
\label{eq:section-3-2-2}
    \mathbf{\hat G}_k = \arg\min_{\mathbf{G}_k} \| \mathbf{Y}_{k}-\mathbf{\Phi}_{k}\mathbf{G}_k\|^2_\text{F} + \mathbf{\gamma} \| \mathbf{G}_k \|^{2}_\text{F},
\end{aligned}
\end{equation}
where $\gamma$ is the regularization parameter and $\| \cdot \|_\text{F}^2$ is the Frobenius norm.
The analytical solution to \eqref{eq:section-3-2-2} can be readily derived using the method of least squares as:
\begin{equation}
\label{eq:section-3-2-3}
  \mathbf{\hat G}_k =(\mathbf{\Phi}_{k}^\top \mathbf{\Phi}_{k}+\gamma\mathbf{I})^{-1} \mathbf{\Phi}_{k}^\top\mathbf{Y}_{k}.
\end{equation}

Concurrently, the client $k$ computes its \textit{Auto-correlation Matrix} according to \eqref{eq:section-3-2-4} as well, and uploads both this matrix $\mathbf{A}_k$ and its local primary stream $\mathbf{\hat G}_k$ to the server.
\begin{equation}
\label{eq:section-3-2-4}
  \mathbf{A}_k = \mathbf{\Phi}_{k}^\top \mathbf{\Phi}_{k}+\gamma\mathbf{I}.
\end{equation}

Upon receiving $\{\mathbf{A}_k\}_{k=1}^{K}$ and $\{\mathbf{\hat G}_k\}_{k=1}^{K}$ from the clients, the server can analytically aggregate all the clients' local knowledge, thereby producing the global primary stream $\mathbf{\hat G}$.
Specifically, the server first updates the \textit{Aggregated Auto-correlation Matrix} as follows:
\begin{equation}
\label{eq:section-3-2-5}
\mathbf{\tilde A}_{k} = \mathbf{\tilde A}_{k-1} + \mathbf{A}_k = \sum\nolimits_{i=1}^k \mathbf{A}_i.
\end{equation}

Subsequently, the server constructs the \textit{Fused Knowledge Matrix}, which is initialized with $\mathbf{F}_{1}=\mathbf{\hat P}_1$ and recursively updated to incorporate each client's local knowledge:
\begin{equation}
\label{eq:section-3-2-6}
\mathbf{F}_{k} = \mathbf{\Lambda}_{k} \mathbf{F}_{k-1} + \mathbf{\Delta}_{k} \mathbf{\hat G}_k,
\end{equation}
where
\begin{equation}
\label{eq:section-3-2-7}
\begin{cases}
\mathbf{\Lambda}_{k} = \mathbf{I} - (\mathbf{\tilde A}_{k-1})^{-1} \mathbf{A}_k (\mathbf{I} - (\mathbf{\tilde A}_{k})^{-1} \mathbf{A}_k), \\
\mathbf{\Delta}_{k} = \mathbf{I} - (\mathbf{A}_k)^{-1} \mathbf{\tilde A}_{k-1} (\mathbf{I} - (\mathbf{\tilde A}_{k})^{-1} \mathbf{\tilde A}_{k-1}).
\end{cases}
\end{equation}

Once the knowledge from all $K$ clients has been integrated into the final \textit{Fused Knowledge Matrix} $\mathbf{F}_{K}$, the server further computes the global primary stream $\mathbf{\hat G}$ as follows:
\begin{equation}
\label{eq:section-3-2-8}
    \mathbf{\hat G}=[\mathbf{\tilde A}_{K} - (K-1)\gamma \mathbf{I}]^{-1} \mathbf{\tilde A}_{K} \mathbf{F}_{K}.
\end{equation}

It is worth noting that the resulting $\mathbf{\hat G}$ coincides exactly with the optimal solution defined in \eqref{eq:objective-2}, and the detailed proof is provided in \textbf{Theorem 1}.
More encouragingly, the final result $\mathbf{\hat G}$ is also independent of the order of clients. 
Consequently, the server can efficiently embed each client's knowledge into the \textit{Fused Knowledge Matrix} upon arrival, rather than adhering to a strict aggregation sequence.
Finally, the server distributes the primary stream $\mathbf{\hat G}$ to all clients.

\subsection{Refinement Stream for Local Personalization}
After receiving the primary stream $\mathbf{\hat G}$, each client $k$ leverages its local data to construct the corresponding refinement stream for local personalization.
Analogously, we also derive the closed-form solution of \eqref{eq:objective-1} to serve as each client's individual refinement stream, further correcting the prediction bias of the primary stream on the local dataset.

Specifically, each client $k$ first extracts its refinement-stream feature matrix $\mathbf{\Psi}_{k}$ using a distinct random projection and activation function to enhance the linear separability:
% , as follows:
\begin{equation}
    \mathbf{\Psi}_{k} = \sigma_\text{R} (\text{Backbone}(\mathbf{X}_k, \Theta) \mathbf{R}_\text{{R}}). \label{eq:section-3-3-1}
\end{equation}

Here, $\mathbf{\Psi}_{k}\in \mathbb{R}^{N_k \times d_\text{R}}$ denotes the feature matrix for the client $k$'s refinement stream, while $\sigma_\text{R}$ and $\mathbf{R}_\text{{R}}$ are the corresponding activation function and random projection matrix.

Subsequently, the client $k$ utilizes the obtained global primary stream $\mathbf{\hat G}$, the activated feature matrices $\mathbf{\Phi}_{k}$ and $\mathbf{\Psi}_{k}$, and the label matrix $\mathbf{Y}_{k}$ to construct its individual refinement stream $\mathbf{\hat P}_k$.
The overall optimization objective in \eqref{eq:objective-1} can then be formulated locally for each client as:
\begin{equation}
\begin{aligned}
\label{eq:section-3-3-2}
    \mathbf{\hat P}_k = \arg\min_{\mathbf{P}_k} \|(\mathbf{Y}_{k}-\mathbf{\Phi}_{k}\mathbf{\hat G})-\mathbf{\Psi}_{k}\mathbf{P}_k\|^2+{\color{black}\mathbf{\beta} \| \mathbf{P}_k \|^{2}},
\end{aligned}
\end{equation}

Here, $\beta$ denotes the regularization parameter for the refinement stream.
Similarly, by applying the least squares method, the closed-form solution of \eqref{eq:section-3-3-2} can be expressed as follows, with detailed proof provided in \textbf{Theorem 2}.
\begin{equation}
\label{eq:section-3-3-3}
  \mathbf{\hat P}_k =(\mathbf{\Psi}_{k}^\top \mathbf{\Psi}_{k}+\beta\mathbf{I})^{-1} (\mathbf{\Psi}_{k}^\top\mathbf{Y}_{k}-\mathbf{\Psi}_{k}^\top\mathbf{\Phi}_{k}\mathbf{\hat G}).
\end{equation}

Thus, the primary stream $\mathbf{\hat G}$ and the refinement stream $\mathbf{\hat P}_k$ are combined to construct the final model for each client $k$.
Consequently, during inference, for any input $\mathbf{\hat X}_k$, the corresponding inference result $\mathbf{\hat Y}_k$ is computed as:
\begin{equation}
\label{eq:section-3-3-4}
  \mathbf{\hat Y}_k = \mathbf{\hat \Phi}_{k} \mathbf{\hat G} + \lambda \mathbf{\hat \Psi}_{k} \mathbf{\hat P}_k,
\end{equation}
where $\mathbf{\hat \Phi}_{k}$ and $\mathbf{\hat \Psi}_{k}$ are the activated feature matrices obtained from $\mathbf{\hat X}_k$ via \eqref{eq:section-3-2-1} and \eqref{eq:section-3-3-1}, respectively.
The balance hyperparameter $\lambda$ controls the trade-off between generalization and personalization, where a larger $\lambda$ places greater emphasis on personalization.
For clarity, we summarize the detailed procedures of our APFL in \textbf{Algorithm 1} within the appendix.

\subsection{Theoretical Analyses}
\label{Section-3.4}
Here, we theoretically analyze the validity, privacy, and efficiency of our APFL, respectively.
Specifically, to establish its validity, we first demonstrate that the primary and refinement streams derived from our APFL are precisely equivalent to the closed-form solutions to \eqref{eq:objective-2} and \eqref{eq:objective-1}, and further prove APFL's ideal property of heterogeneity invariance.

\noindent\textbf{Theorem 1:} 
The primary stream $\mathbf{\hat G}$ derived from our APFL is exactly equivalent to the optimal solution that is obtained through empirical risk minimization centrally over the complete dataset $D_{1:K}$, as defined by the objective \eqref{eq:objective-2}.

\noindent \textbf{\textit{Proof.}} See Appendix A.

\noindent\textbf{Theorem 2:} The refinement streams $\{ \mathbf{\hat P}_k \}_{k=1}^K$ derived from our APFL are exactly equivalent to the optimal solutions obtained through empirical risk minimization over the local datasets $\{ D_k \}_{k=1}^K$, as defined by the objective \eqref{eq:objective-1}.

\noindent \textbf{\textit{Proof.}} See Appendix A.

\noindent\textbf{Theorem 3 (Heterogeneity Invariance):} For any client $k$, the resulting final model $\{ \mathbf{\hat G},  \mathbf{\hat P}_k\}$ derived from our APFL are independent of the data distributions of other clients, depending solely on the complete dataset $D_{1:K}$ and its local dataset $D_k$.
Formally, given two arbitrary PFL systems with distinct data distributions $\{ D_i \}_{i=1}^K$ and $\{ D'_i \}_{i=1}^K$.
As long as $D_k = D'_k$ and $\bigcup_{i=1}^K D_i =  \bigcup_{i=1}^K D'_i$, their resulting $\{ \mathbf{\hat G},  \mathbf{\hat P}_k\}$ will be identical.

\noindent \textbf{\textit{Proof.}} See Appendix A.

Second, we analyze the privacy of our proposed APFL, proving that it is impossible to fully reconstruct the clients' private information from their raw data, as follows.

\noindent\textbf{Theorem 4 (Privacy):} It is impossible to infer any client $k$'s private information $\mathbf{\Phi}_{k}$, $\mathbf{\Psi}_{k}$, and $\mathbf{Y}_{k}$, based on their submitted local knowledge $\mathbf{A}_k$ and $\mathbf{\hat G}_k$ within our APFL.

\noindent \textbf{\textit{Proof.}} See Appendix B.

Third, we further analyze the efficiency of our APFL.
Specifically, the complexities of computation and communication for each client are $O({d_\text{P}}^3+{d_\text{R}}^3 + N_k{d_\text{P}}^2 + N_k{d_\text{R}}^2)$ and $O({d_\text{P}}^2)$, while those for the server are $O(K{d_\text{P}}^3)$ and $O(cK{d_\text{P}})$. 
For brevity, we present only the final results here, and the detailed derivation is provided in Appendix C.

%% file: sec/4_experiments.tex
{
\fontsize{9}{11}\selectfont
\begin{table*}[t]
\centering
\caption{
The overall results of the comparison.
The best result is highlighted in \textbf{bold}, and the second-best result is \underline{underlined}.
The ``Advance \textbf{$\uparrow$}” refers to the performance advantage of our proposed APFL compared to the second-best result.
% Furthermore, the superscript $^{*}$ denotes that the advance is statistically significant, based on a T-test at a level of 0.05.
} 
\renewcommand{\arraystretch}{1.16}
\setlength{\tabcolsep}{1mm}
\begin{NiceTabular}{@{}l|cccccc|cccccc@{}}
\toprule
% \multicolumn{1}{c|}{\multirow{2}{*}{Baseline}} &
\multicolumn{1}{c}{\multirow{4}{*}{{Baseline}}} &
  \multicolumn{6}{c}{CIFAR-100} &
  \multicolumn{6}{c}{ImageNet-R} \\ \cmidrule(l){2-13} 
  \multicolumn{1}{c}{} &
  \multicolumn{3}{c|}{50 Clients} &
  \multicolumn{3}{c}{100 Clients} &
  \multicolumn{3}{c|}{50 Clients} &
  \multicolumn{3}{c}{100 Clients} \\ \cmidrule(l){2-13} 
\multicolumn{1}{c}{} &
  $\alpha=0.1$ &
  $\alpha=0.5$ &
  \multicolumn{1}{c|}{$\alpha=1.0$} &
  $\alpha=0.1$ &
  $\alpha=0.5$ &
  \multicolumn{1}{c}{$\alpha=1.0$} &
  $\alpha=0.1$ &
  $\alpha=0.5$ &
  \multicolumn{1}{c|}{$\alpha=1.0$} &
  $\alpha=0.1$ &
  $\alpha=0.5$ &
  $\alpha=1.0$ \\ \midrule
FedAvg & 43.53\% & 47.75\% & \multicolumn{1}{c|}{48.91\%} & 39.28\% & 43.31\% & \multicolumn{1}{c|}{43.28\%} & 6.30\% & 7.14\% & \multicolumn{1}{c|}{7.61\%} & 4.90\% & 4.60\% & 5.00\% \\
$\text{FedAvg}^\text{FT}$ & 58.81\% & 41.38\% & \multicolumn{1}{c|}{37.73\%} & 49.05\% & 31.37\% & \multicolumn{1}{c|}{26.91\%} & 20.42\% & 8.14\% & \multicolumn{1}{c|}{6.15\%} & 18.68\% & 7.52\% & 4.84\% \\
FedProx & 42.74\% & 46.73\% & \multicolumn{1}{c|}{48.19\%} & 38.37\% & 42.38\% & \multicolumn{1}{c|}{42.54\%} & 6.28\% & 7.36\% & \multicolumn{1}{c|}{7.21\%} & 4.85\% & 4.50\% & 4.99\% \\
$\text{FedProx}^\text{FT}$ & 57.63\% & 41.32\% & \multicolumn{1}{c|}{37.73\%} & 48.64\% & 31.38\% & \multicolumn{1}{c|}{26.91\%} & 20.40\% & 8.11\% & \multicolumn{1}{c|}{6.13\%} & 18.70\% & 7.53\% & 4.82\% \\
Ditto & 69.80\% & 50.18\% & \multicolumn{1}{c|}{43.72\%} & 66.80\% & 41.32\% & \multicolumn{1}{c|}{33.46\%} & 25.35\% & 8.57\% & \multicolumn{1}{c|}{6.05\%} & 22.85\% & 7.88\% & 5.10\% \\
FedALA & \underline{72.89\%} & {52.60\%} & \multicolumn{1}{c|}{46.42\%} & \underline{70.63\%} & {47.11\%} & \multicolumn{1}{c|}{38.20\%} & 32.01\% & 12.29\% & \multicolumn{1}{c|}{8.86\%} & \underline{28.20\%} & 10.15\% & 5.91\% \\
FedDBE & 48.52\% & 49.96\% & \multicolumn{1}{c|}{{50.11\%}} & 43.20\% & 44.58\% & \multicolumn{1}{c|}{{44.41\%}} & 9.20\% & 8.46\% & \multicolumn{1}{c|}{8.31\%} & 7.40\% & 5.61\% & 5.78\% \\
FedAS & 56.70\% & 41.37\% & \multicolumn{1}{c|}{37.78\%} & 47.79\% & 30.73\% & \multicolumn{1}{c|}{27.28\%} & 20.32\% & 7.96\% & \multicolumn{1}{c|}{6.00\%} & 18.33\% & 7.37\% & 4.65\% \\
FedPCL & 23.99\% & 11.33\% & \multicolumn{1}{c|}{12.52\%} & 25.83\% & 11.82\% & \multicolumn{1}{c|}{11.13\%} & 0.83\% & 0.88\% & \multicolumn{1}{c|}{1.13\%} & 1.72\% & 1.11\% & 1.23\% \\
FedSelect & 72.46\% & 49.44\% & \multicolumn{1}{c|}{41.58\%} & 68.44\% & 42.71\% & \multicolumn{1}{c|}{34.41\%} & \underline{34.83\%} & {14.94\%} & \multicolumn{1}{c|}{{9.59\%}} & 28.11\% & {11.01\%} & {7.09\%} \\
AFL         & 58.99\% & \underline{58.25\%} & \multicolumn{1}{c|}{\underline{54.97\%}} & 56.96\% & \underline{58.52\%} & \multicolumn{1}{c}{\underline{55.16\%}} & 32.87\% & \underline{33.85\%} & \multicolumn{1}{c|}{\underline{32.94\%}} & 28.09\% & \underline{26.66\%} & \underline{25.03\%} \\
\midrule
APFL         & \textbf{80.72\%} & \textbf{67.34\%} & \multicolumn{1}{c|}{\textbf{59.72\%}} & \textbf{77.08\%} & \textbf{64.64\%} & \multicolumn{1}{c}{\textbf{58.38\%}} & \textbf{49.92\%} & \textbf{37.84\%} & \multicolumn{1}{c|}{\textbf{34.04\%}} & \textbf{43.65\%} & \textbf{29.40\%} & \textbf{26.62\%} \\
\rowcolor{gray!15}
\textbf{Advance \textbf{$\uparrow$}} & \textbf{7.83\%$^{*}$} & \textbf{9.09\%$^{*}$} & \multicolumn{1}{c|}{\textbf{4.75\%$^{*}$}} & \textbf{6.45\%$^{*}$} & \textbf{6.12\%$^{*}$} & \multicolumn{1}{c}{\textbf{3.22\%$^{*}$}} & \textbf{15.09\%$^{*}$} & \textbf{3.99\%$^{*}$} & \multicolumn{1}{c|}{\textbf{1.10\%$^{*}$}} & \textbf{15.45\%$^{*}$} & \textbf{2.74\%$^{*}$} & \textbf{1.59\%$^{*}$} \\
\bottomrule
\end{NiceTabular}
% }
\label{table:overall}
\end{table*}
}

\section{Experiments}
\label{sec:Experiments}

\subsection{Experiment Setting}
\label{sec:Experimental Setup}

\textbf{Datasets \& Settings.} 
To comprehensively evaluate the performance of our APFL, we conduct extensive experiments on 2 benchmark datasets: CIFAR-100 \cite{dataset_CIFAR} and ImageNet-R \cite{dataset_ImageNet-R}.
Moreover, to simulate different non-IID scenarios in PFL, we partition the data among 50 and 100 clients using the Dirichlet distribution \cite{Dirichlet} with the concentration parameter $\alpha \in \{ 0.1, 0.5, 1.0 \}$.
Notably, a larger number of clients (e.g., 100 clients) and a smaller concentration parameter (e.g., $\alpha = 0.1$) indicate more severe data heterogeneity, as a more challenging non-IID scenario.
More details on the experimental configurations can be found in Appendix D.

\textbf{Baselines \& Metrics.} 
We compare our proposed APFL against both classical FL baselines and state-of-the-art PFL baselines.
The FL ones include FedAvg \cite{FedAvg}, FedProx \cite{FedProx}, their fine-tuned variants (i.e., $\text{FedAvg}^\text{FT}$ \& $\text{FedProx}^\text{FT}$), as well as the analytic learning-based approach AFL \cite{AFL}.
The PFL ones include Ditto \cite{Ditto}, FedALA \cite{FedALA}, FedAS \cite{FedAS}, FedPCL \cite{FedPCL}, FedDBE \cite{FedDBE}, and FedSelect \cite{FedSelect}.
For fairness, all methods employ the same open-source backbone of ViT-MAE-Base for feature extraction~\cite{ViT-MAE}.
All gradient-based baselines are configured with 3 local training epochs and 200 global communication rounds.
We establish the average accuracy of each client's final model on its respective local test set as the core metric for evaluating the performance of all methods.
Running time and communication overhead are utilized as metrics to evaluate the efficiency.

\begin{figure*}[!h]
  \centering
  % 第一行
  \subfloat[Accuracy vs. Aggregation Round]{\includegraphics[width=0.6666\columnwidth]{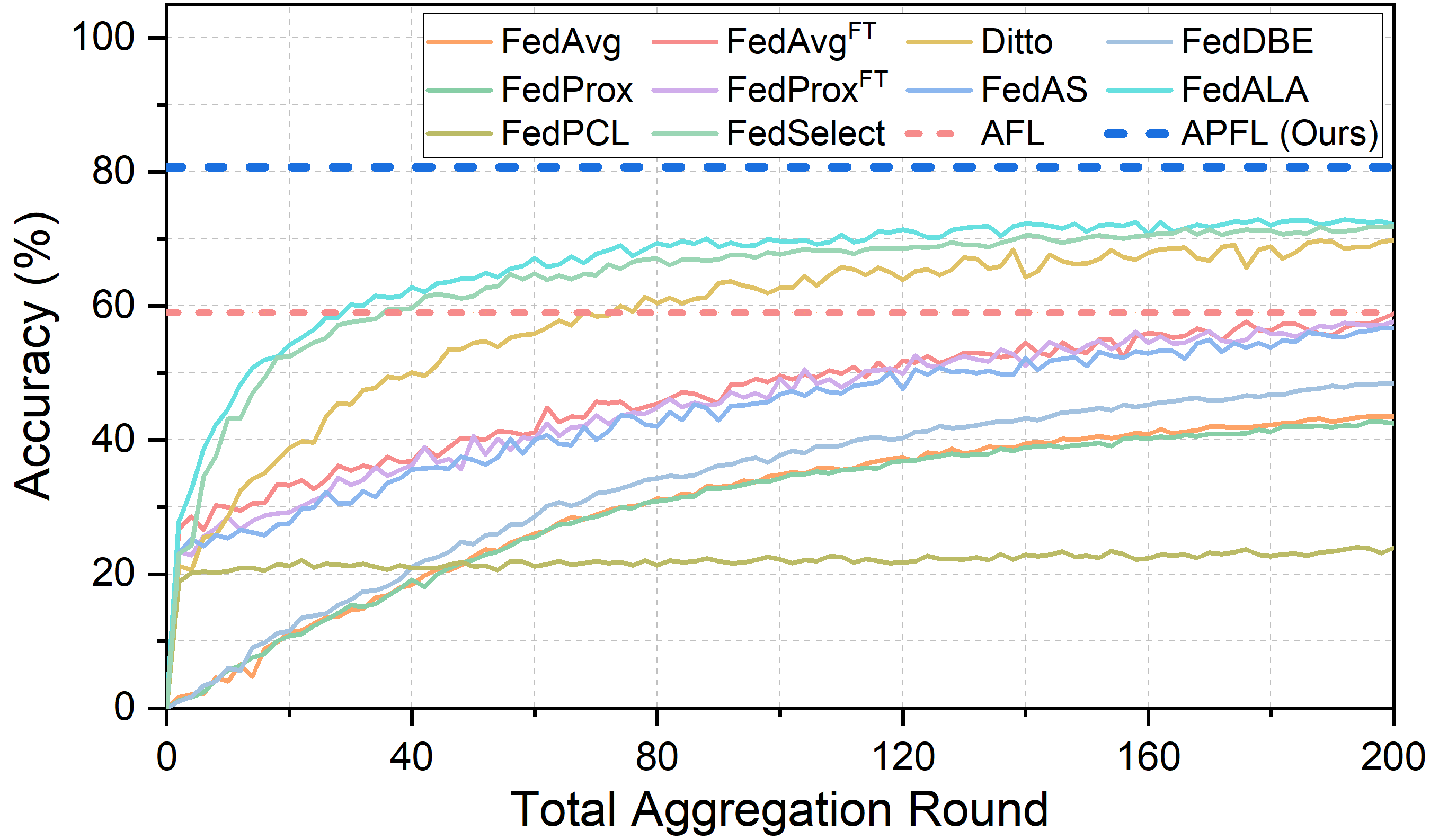}
  \label{fig:exp1a}}%
  \hfill
  \subfloat[Accuracy vs. Computation Overhead]{\includegraphics[width=0.6666\columnwidth]{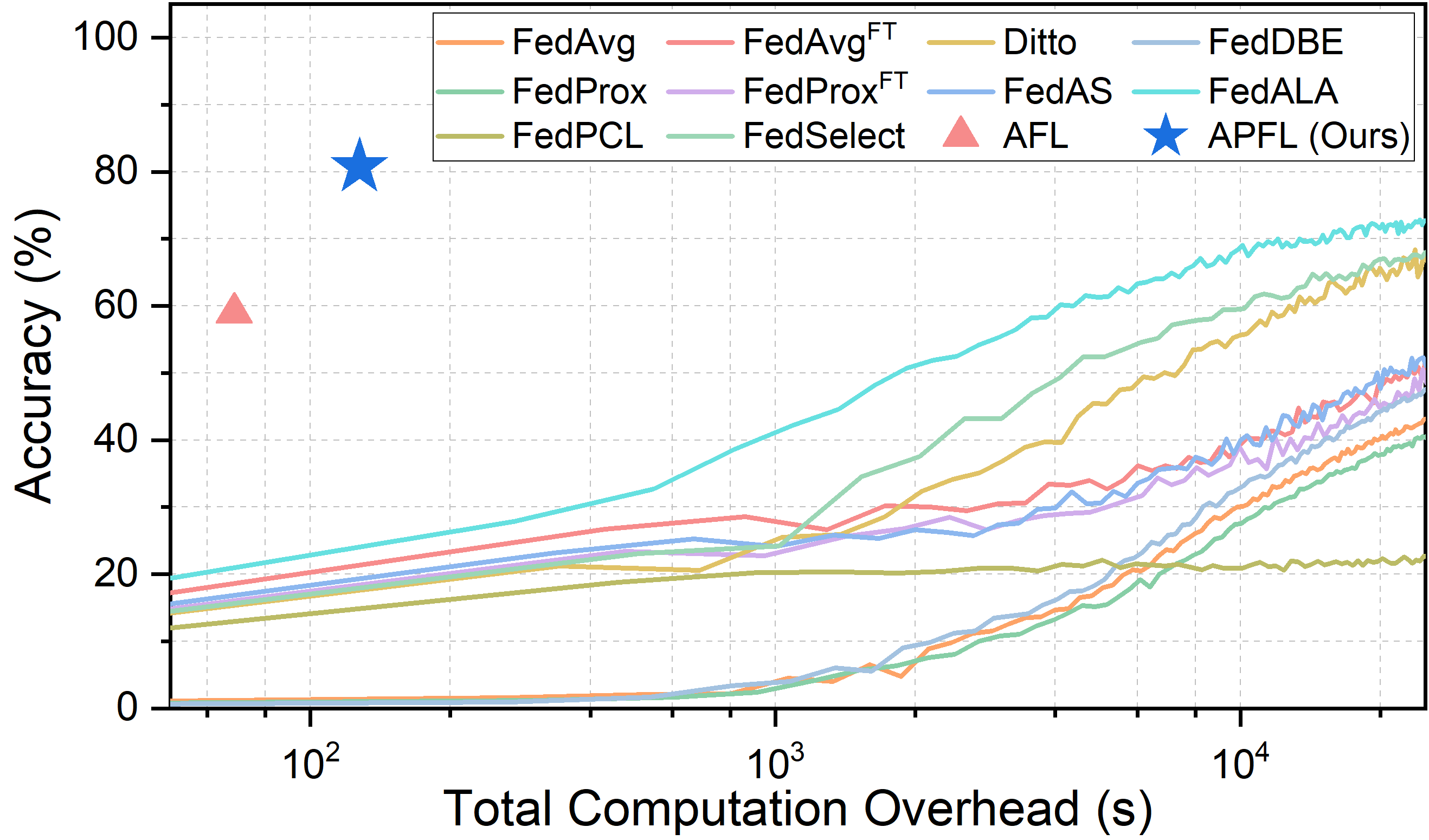}
  \label{fig:exp1b}}%
  \hfill
  \subfloat[Accuracy vs. Communication Overhead]{\includegraphics[width=0.6666\columnwidth]{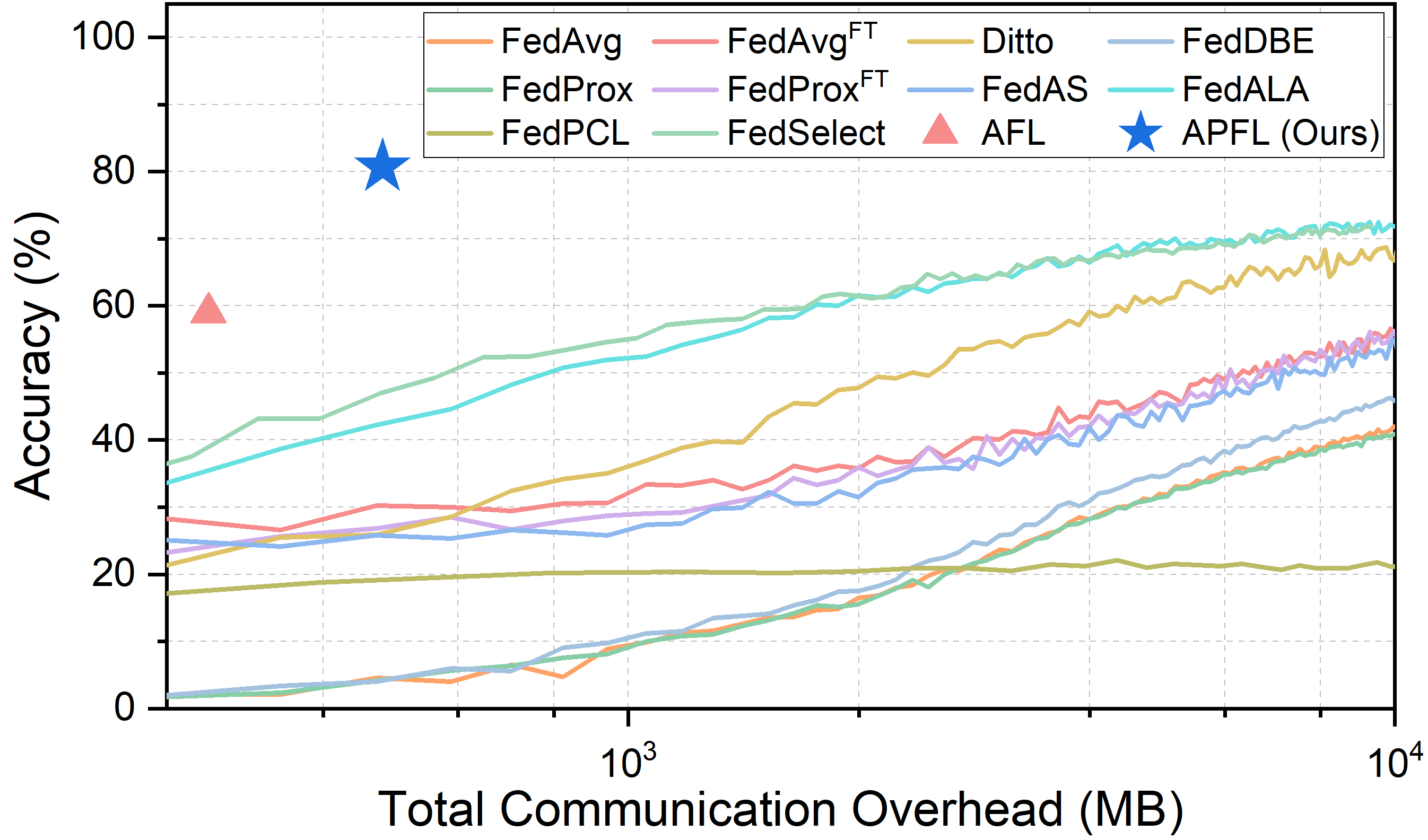}
  \label{fig:exp1c}}%
  \hfill
  \caption{Efficiency evaluations on the CIFAR-100 dataset.}
  \label{fig:exp1}
\end{figure*}

\begin{figure*}[!h]
  \centering
  % 第一行
  \subfloat[Accuracy vs. Aggregation Round]{\includegraphics[width=0.6666\columnwidth]{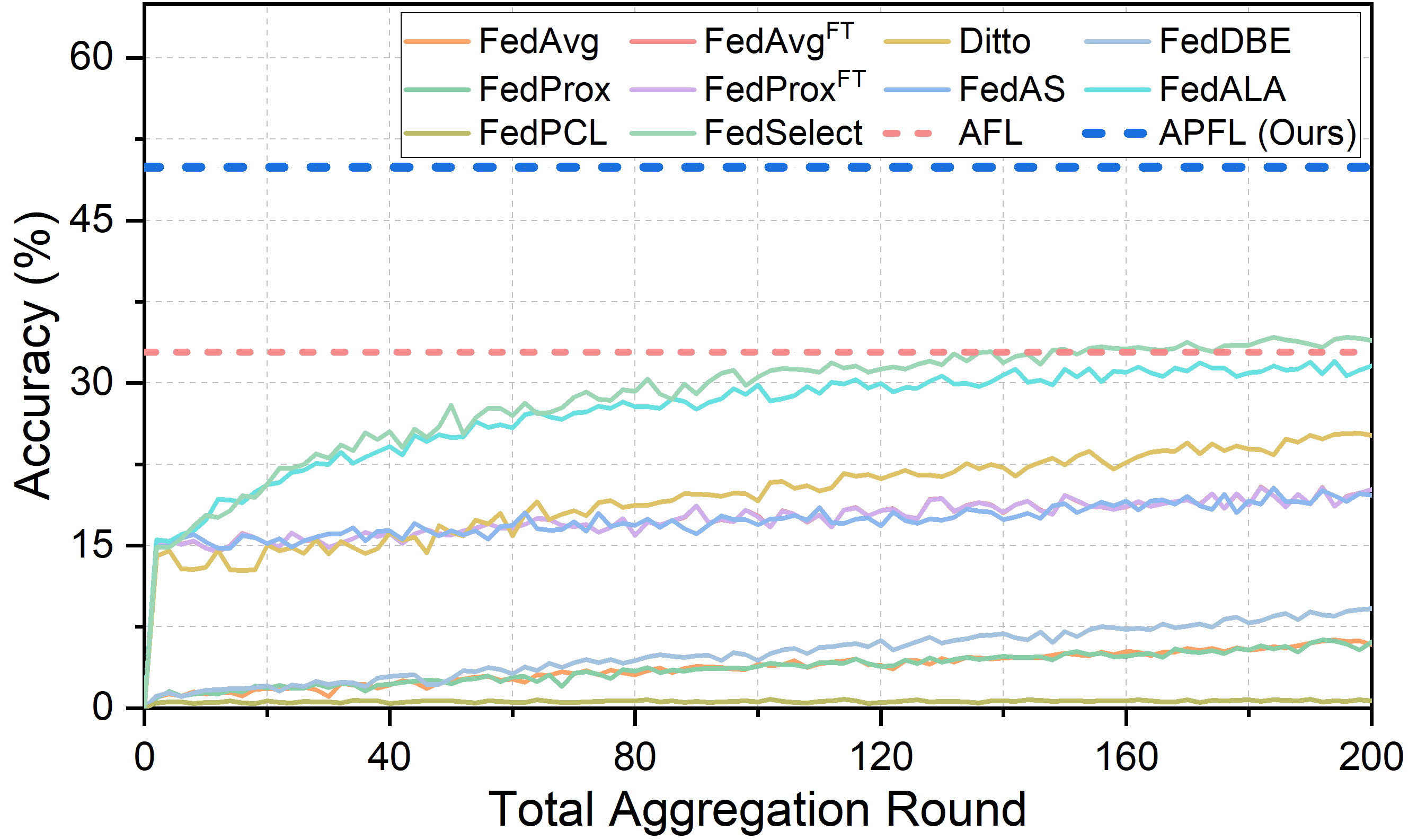}
  \label{fig:exp2a}}%
  \hfill
  \subfloat[Accuracy vs. Computation Overhead]{\includegraphics[width=0.6666\columnwidth]{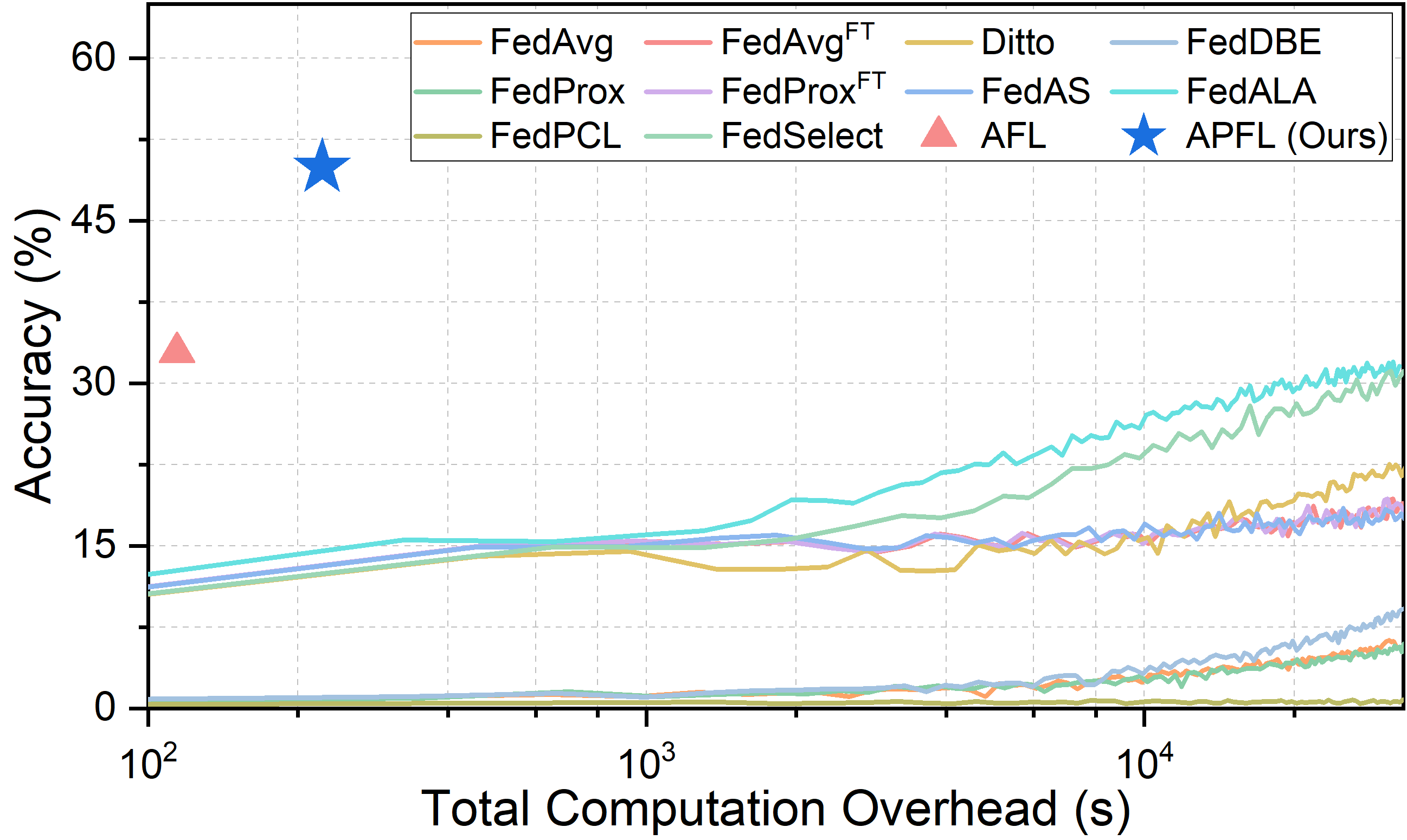}
  \label{fig:exp2b}}%
  \hfill
  \subfloat[Accuracy vs. Communication Overhead]{\includegraphics[width=0.6666\columnwidth]{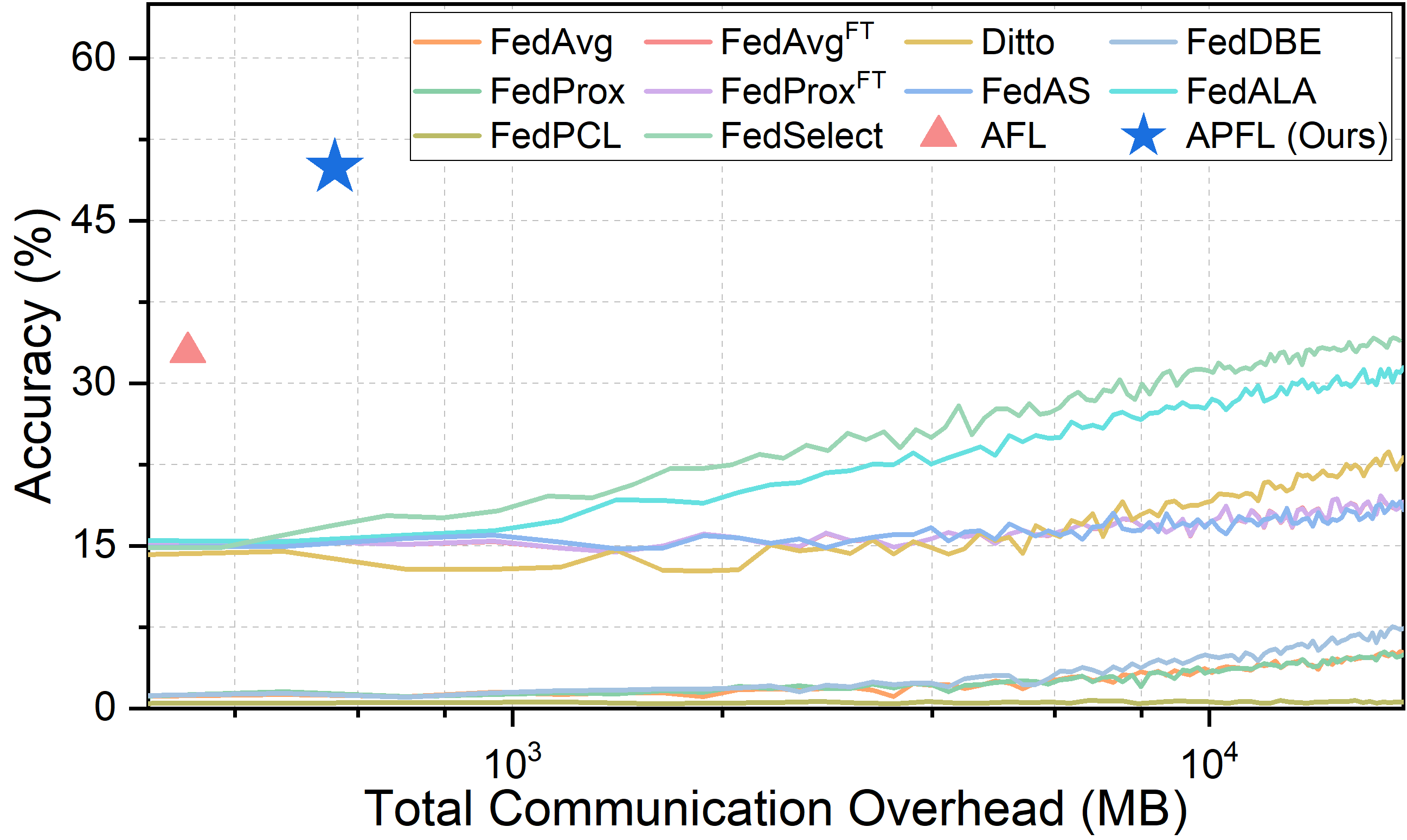}
  \label{fig:exp2c}}%
  \hfill
  \caption{Efficiency evaluations on the ImageNet-R dataset.}
  \label{fig:exp2}
  \vspace{-0.05cm}
\end{figure*}

{
\fontsize{9}{11}\selectfont
\begin{table*}[ht]
\centering
\caption{Accuracy of our APFL and its individual streams under varying balance hyperparameter $\lambda$.} 
\renewcommand{\arraystretch}{1.2}
\setlength{\tabcolsep}{1.5mm}
\begin{NiceTabular}{@{}c|c|ccccc|ccccc@{}}
\toprule
\multirow{2.5}{*}{\textbf{Setting}} &
\multirow{2.5}{*}{\textbf{Stream}} & \multicolumn{5}{c|}{CIFAR-100} & \multicolumn{5}{c}{ImageNet-R} 
\\ \cmidrule(l){3-12} & & $\lambda= 0.1$ & $\lambda= 0.3$ & $\lambda= 0.5$ & $\lambda= 0.7$ & $\lambda= 0.9$ & $\lambda= 0.1$ & $\lambda= 0.3$ & $\lambda= 0.5$ & $\lambda= 0.7$ & $\lambda= 0.9$ \\
\midrule
\multirow{2}{*}{$\alpha=0.1$}
& Primary & 56.41\% & 56.41\% & 56.41\% & 56.41\% & 56.41\% & 33.37\% & 33.37\% & 33.37\% & 33.37\% & 33.37\% \\
% Refinement-only & 11.13\% & 11.13\% & 11.13\% & 11.13\% & 11.13\% & 11.13\% & 11.13\% & 11.13\% & 11.13\% & 11.13\% \\
& Dual (Ours) & 69.33\% & 78.90\% & \textbf{79.20\%} & 78.09\% & 76.76\% & 47.04\% & \textbf{49.92\%} & 45.96\% & 42.91\% & 41.18\%\\
\midrule
\multirow{2}{*}{$\alpha=0.5$}
& Primary & 55.20\% & 55.20\% & 55.20\% & 55.20\% & 55.20\% & 33.85\% & 33.85\% & 33.85\% & 33.85\% & 33.85\% \\
% Refinement-only & 11.13\% & 11.13\% & 11.13\% & 11.13\% & 11.13\% & 11.13\% & 11.13\% & 11.13\% & 11.13\% & 11.13\% \\
& Dual (Ours) & 59.60\% & 64.51\% & \textbf{64.60\%} & 62.82\% & 60.84\% & \textbf{36.82\%} & 36.21\% & 30.98\% & 27.13\% & 24.36\%\\
\midrule
\multirow{2}{*}{$\alpha=1.0$}
& Primary & 54.97\% & 54.97\% & 54.97\% & 54.97\% & 54.97\% & 32.94\% & 32.94\% & 32.94\% & 32.94\% & 32.94\% \\
% Refinement-only & 11.13\% & 11.13\% & 11.13\% & 11.13\% & 11.13\% & 11.13\% & 11.13\% & 11.13\% & 11.13\% & 11.13\% \\
& Dual (Ours) & 57.52\% & \textbf{59.72\%} & 59.01\% & 56.93\% & 54.86\% & \textbf{34.04\%} & 33.10\% & 27.25\% & 22.58\% & 19.91\%\\
\bottomrule
\end{NiceTabular}
\label{table:2}
\end{table*}
}

\subsection{Overall Results}
As detailed in Table~\ref{table:overall}, we compared the performance of our APFL with other baselines across various datasets and diverse experimental settings.
Overall, our APFL consistently achieves the best performance across all experimental settings, outperforming the state-of-the-art baseline with significant advances ranging from 1.10\% to 15.45\%.
Notably, our APFL consistently outperforms AFL across all experimental settings.
Furthermore, since AFL does not account for personalization, it ceases to be the strongest baseline in highly non-IID scenarios (i.e., $\alpha=0.1$), where FedALA and FedSelect emerge as the top performers among baselines.
Correspondingly, our APFL's performance advantage over AFL further expands with increasing data heterogeneity.
For gradient-based baselines, the increasing data heterogeneity detrimentally affects model aggregation, resulting in a widespread performance degradation for FL methods, such as FedAvg and FedProx. 
Conversely, the increasing data heterogeneity also makes intra-client data distributions more localized and easier to personalize, thereby enhancing the performance of PFL baselines.

Concurrently, it is observed that as the number of clients increases (i.e., from 50 to 100 clients), the performance of all methods generally declines.
This evidence is because a larger number of clients not only detrimentally affects model aggregation but also results in each client's local dataset becoming sparser, thus making effective personalization more difficult.
Furthermore, despite our extensive hyperparameter tuning efforts, the average accuracy of most gradient-based baselines on the challenging ImageNet-R dataset still languished below 10.00\%, thereby highlighting the inherent limitations of the existing gradient-based methods.

\subsection{Efficiency Evaluations}

Subsequently, we conducted comprehensive efficiency comparisons between our APFL and other baselines.
Here, we set $\alpha$ to 0.1 and the number of clients to 50, performing experiments on both the CIFAR-100 and ImageNet-R datasets. 
The corresponding experimental results for these two datasets are shown in Figure~\ref{fig:exp1} and Figure~\ref{fig:exp2}, respectively.
Each figure contains three subfigures, depicting the average accuracy of all methods across aggregation rounds, computation overhead, and communication overhead.

We first focus on the efficiency comparison results on the CIFAR-100 dataset.
As shown in Figure~\ref{fig:exp1}(a), all gradient-based baselines rely on multiple rounds of model aggregation, with their performance gradually improving over these rounds.
Conversely, APFL remarkably achieves the best performance with only a single aggregation round per client, decisively outperforming the baselines' final results even after $200$ rounds of aggregation.
Subsequently, Figures~\ref{fig:exp1}(b) and~\ref{fig:exp1}(c) indicate our APFL's efficiency advantages in terms of computation and communication overhead.
Specifically, our APFL, represented by a blue star, is conspicuously positioned in the top-left corner of the plots, signifying the best performance achieved with very low overhead.
Quantitatively, compared to AFL, our proposed APFL substantially improved performance by up to 21.73\%, incurring only little additional computation and communication overhead.
In addition, compared to these gradient-based baselines, our proposed APFL achieves at least 7.83\% higher performance while incurring less than 1.00\% of their computation overhead and 6.00\% of their communication overhead.

Figure~\ref{fig:exp2} depicts the efficiency comparison results for the ImageNet-R dataset, which are highly consistent with those presented in Figure~\ref{fig:exp1}.
Notably, despite extensive hyperparameter tuning efforts and the deployment of their most favorable setting (i.e., $\alpha=0.1$ and 50 clients), the baselines consistently demonstrated weak performance on the ImageNet-R dataset, making the performance advantage of APFL even more pronounced than that on the CIFAR-100 dataset.
This evidence further underscores the advantages of our APFL on challenging datasets, as it fundamentally addresses the issue of non-IID data plaguing gradient-based baselines, and its dual analytic streams also achieve both collective generalization and individual personalization in PFL.

\vspace{-0.03cm}

To further demonstrate the efficiency advantages of our APFL, we present the overheads of our APFL and baselines across varying aggregation rounds.
The detailed results are depicted in Figures 5-6 within Appendix D.
As the gradient-based baselines rely on multiple rounds of model aggregation and incur similar overheads, we select the popular FL baseline, FedAvg, and the PFL baseline, FedPCL, for comparison.
As evident, our APFL relies solely on a single round of model aggregation for each client, rendering its overheads independent of the aggregation rounds.
In contrast, the baselines necessitate multiple rounds of aggregation to gradually improve their model performance.
Specifically, our APFL's total overhead is at least 93.00\% and 59.27\% lower than that of the baselines even within the initial 20 rounds, and it achieves a notable reduction of up to 99.30\% and 95.93\% compared to their cumulative overhead over 200 rounds, on computation and communication overheads, respectively.
All these results show the superior efficiency of our APFL.

\subsection{Sensitivity Analyses}

Last but not least, we conducted comprehensive sensitivity analyses to examine the effectiveness of our dual-stream analytic models, as well as the impact of various hyperparameters and nonlinear activation functions.
To ensure clarity, our APFL is designated as ``Dual", while the variant leveraging solely the primary stream is termed ``Primary".

First, let's focus on the balance hyperparameter $\lambda$, where a larger $\lambda$ increases the weight of the refinement stream and places greater emphasis on personalization.
As shown in Table~\ref{table:2}, our APFL consistently outperforms the variant relying solely on the primary stream, highlighting the effectiveness of our dual analytic streams.
Notably, APFL’s performance follows a rising-then-falling trend as $\lambda$ increases.
This trend arises because excessively small $\lambda$ results in insufficient personalization, while too large $\lambda$ risks overfitting to local data. Thus, achieving optimal performance necessitates a careful trade-off between personalization and generalization.
Specifically, the optimal $\lambda$ lies between 0.3 and 0.5 for CIFAR-100, and between 0.1 and 0.3 for ImageNet-R.
Furthermore, as $\alpha$ increases (i.e., data heterogeneity diminishes), the optimal $\lambda$ gradually decreases, owing to the diminished personalization inherent in clients' local datasets, which requires less emphasis on personalization.

Second, we analyze the influence of the regularization parameters $\gamma$ and $\beta$.
Specifically, we investigate the individual effects of $\gamma$ and $\beta$ by fixing $\beta=1$ and $\gamma=1$ in turn, while varying the other parameter. 
The comprehensive results are provided in Table 3 and Table 4 within Appendix D.
Our APFL's performance remains stable across a broad range of $\gamma$, with the optimal $\gamma$ found to be 0.01 for CIFAR-100 and ranging from 0.01 to 1 for ImageNet-R.
In contrast, the choice of $\beta$ substantially influences APFL's performance, with optimal $\beta$ between 1 and 10 for CIFAR-100 and fixed at 10 for ImageNet-R.
This disparity in sensitivity stems from the distinct characteristics of the two streams.
The primary stream, which focuses on the complete dataset, is less prone to overfitting and hence insensitive to $\gamma$.
Conversely, the refinement stream, reliant solely on each client's local data with limited volume, is more susceptible to overfitting and thus sensitive to $\beta$.
Furthermore, the optimal value of $\beta$ is notably larger than that of $\gamma$, also attributable to the refinement stream's requirement for greater regularization.

Third, we evaluate the influence of the random projection dimensions $d_\text{P}$ and $d_\text{R}$.
Similarly, we investigate the individual influence of $d_\text{P}$ and $d_\text{R}$ by fixing one at $2^{10}$ while varying the other.
As shown in Tables 5-6 within Appendix D, our APFL's performance exhibits a rising-then-falling trend with increasing $d_\text{P}$ or $d_\text{R}$.
Specifically, the optimal value of $d_\text{P}$ lies within the range of $2^{11}$ and $2^{13}$ for CIFAR-100, and from $2^{10}$ to $2^{11}$ for ImageNet-R.
Meanwhile, the optimal $d_\text{R}$ lies between $2^{10}$ and $2^{11}$ for both CIFAR-100 and ImageNet-R.
Furthermore, while our APFL's performance remains stable with respect to $d_\text{R}$, it is relatively sensitive to $d_\text{P}$, 
as tuning the primary stream impacts both streams' performance.

Fourth, we further investigate the influence of various activation functions, $\sigma_\text{P} (\cdot)$ and $\sigma_\text{R} (\cdot)$, as shown in Tables 7-10 within Appendix D.
Specifically, we analyze the isolated impact of activation functions $\sigma_\text{P} (\cdot)$ and $\sigma_\text{R} (\cdot)$ by fixing one as ReLU and varying the other across eight different activation functions.
As a result, our APFL appears insensitive to different activation functions in both the primary and refinement streams. 
The optimal activation $\sigma_\text{P} (\cdot)$ is observed to be Hardswish or Tanh for CIFAR-100, while it is consistently ReLU for ImageNet-R.
Meanwhile, the optimal activation $\sigma_\text{R} (\cdot)$ is Hardswish or Sigmoid for CIFAR-100, and keeps ReLU for ImageNet-R.
Moreover, the choice of activation function has a relatively greater impact on our APFL's performance on the ImageNet-R dataset than on the CIFAR-100 dataset, likely owing to the former's higher complexity.

\section{Conclusion}
\label{sec:conclusion}

In this paper, we propose APFL to address the challenge of non-IID data in PFL in a gradient-free manner. 
Our APFL uses dual-stream analytic models to achieve both collective generalization and individual personalization. 
Specifically, our APFL integrates a shared primary stream for global generalization across all clients and a dedicated refinement stream for the local personalization of each individual client.
To our knowledge, our APFL represents the first to introduce analytic learning into PFL via dual-stream least squares, handling the non-IID issue while achieving both generalization and personalization.
Theoretical analyses show the validity, privacy, and efficiency of our APFL, particularly its heterogeneity invariance property.
Extensive experiments also validate the superior performance of our APFL.